\documentclass[letterpaper, 10 pt, conference]{IEEEtran}  

\IEEEoverridecommandlockouts                              

\usepackage{multirow}
\usepackage{stfloats} 
\usepackage{graphicx}
\usepackage{booktabs}
\usepackage{amssymb}  

\newcommand{\model}{VLM-E2E}
\newcommand{\nuScenes}{nuScenes}

\title{\LARGE \bf VLM-E2E: Enhancing End-to-End Autonomous Driving with Multimodal Driver Attention Fusion}

\author{Pei Liu, Haipeng Liu, Haichao Liu, Xin Liu, Jinxin Ni, Jun Ma, \textit{Senior Member, IEEE}
\thanks{Pei Liu, Haichao Liu, and Xin Liu are with  The Hong Kong University of Science and Technology (Guangzhou), Guangzhou 511453, China (e-mail: pliu061@connect.hkust-gz.edu.cn; hliu369@connect.hkust-gz.edu.cn; xliu969@connect.hkust-gz.edu.cn).}
\thanks{Haipeng Liu is with Li Auto Inc., Shanghai 201800, China (e-mail: liuhaipeng2012@live.com).}
\thanks{Jinxin Ni is with the School of Aeronautics and Astronautics, Xiamen
University, Xiamen 361102, China (e-mail: nijinxinlxq@outlook.com).}

\thanks{Jun Ma is with The Hong Kong University of Science and Technology (Guangzhou), Guangzhou 511453, China, and also with The Hong Kong University of Science and Technology, Hong Kong SAR, China (e-mail: jun.ma@ust.hk).}
}

\begin{document}
\maketitle
\thispagestyle{empty}
\pagestyle{empty}

\begin{abstract}

Human drivers adeptly navigate complex scenarios by utilizing rich attentional semantics, but the current autonomous systems struggle to replicate this ability, as they often lose critical semantic information when converting 2D observations into 3D space. In this sense, it hinders their effective deployment in dynamic and complex environments.
Leveraging the superior scene understanding and reasoning abilities of Vision-Language Models (VLMs), we propose {\model}, a novel framework that uses the VLMs to enhance training by providing attentional cues. Our method integrates textual representations into Bird's-Eye-View (BEV) features for semantic supervision, which enables the model to learn richer feature representations that explicitly capture the driver's attentional semantics. By focusing on attentional semantics, {\model} better aligns with human-like driving behavior, which is critical for navigating dynamic and complex environments.
Furthermore, we introduce a BEV-Text learnable weighted fusion strategy to address the issue of modality importance imbalance in fusing multimodal information. This approach dynamically balances the contributions of BEV and text features, ensuring that the complementary information from visual and textual modalities is effectively utilized. 
By explicitly addressing the imbalance in multimodal fusion, our method facilitates a more holistic and robust representation of driving environments.
We evaluate {\model} on the {\nuScenes} dataset and achieve significant improvements in perception, prediction, and planning over the baseline end-to-end model, showcasing the effectiveness of our attention-enhanced BEV representation in enabling more accurate and reliable autonomous driving tasks.
\end{abstract}

\section{Introduction}
\label{sec:intro}

Autonomous driving has witnessed remarkable progress in recent years \cite{jiang2023vad, hu2023planning, wang2022detr3d}, with significant advancements in key areas such as perception \cite{li2022bevformer}, motion prediction \cite{gu2023vip3d}, and planning \cite{prakash2021multi}. These developments have laid a solid foundation for achieving more accurate and safer driving decisions. Among these, end-to-end (E2E) autonomous driving has emerged as a transformative paradigm, leveraging large-scale data to demonstrate impressive planning capabilities. By directly mapping raw sensor inputs to driving actions, E2E approaches bypass the need for handcrafted intermediate modules, enabling more flexible and scalable solutions.
However, 
Despite these advancements, traditional end-to-end autonomous driving approaches predominantly predict future trajectories or control signals directly, without explicitly considering the driver’s attention to critical information such as traffic dynamics and navigation cues. E2E systems often struggle in complex and ambiguous scenarios due to their limited ability to reason about high-level semantics and contextual cues, such as traffic rules, driver attention, and dynamic interactions. In contrast, human drivers rely on an attentional decision-making process, where attention to both the surrounding traffic environment and navigation guidance plays a critical role \cite{badre2008cognitive}. For instance, when approaching an intersection, human drivers naturally prioritize traffic signals, pedestrian movements, and lane markings, dynamically adjusting their focus based on the evolving scene.

\begin{figure}
    \centering
    \includegraphics[width=1.0\linewidth]{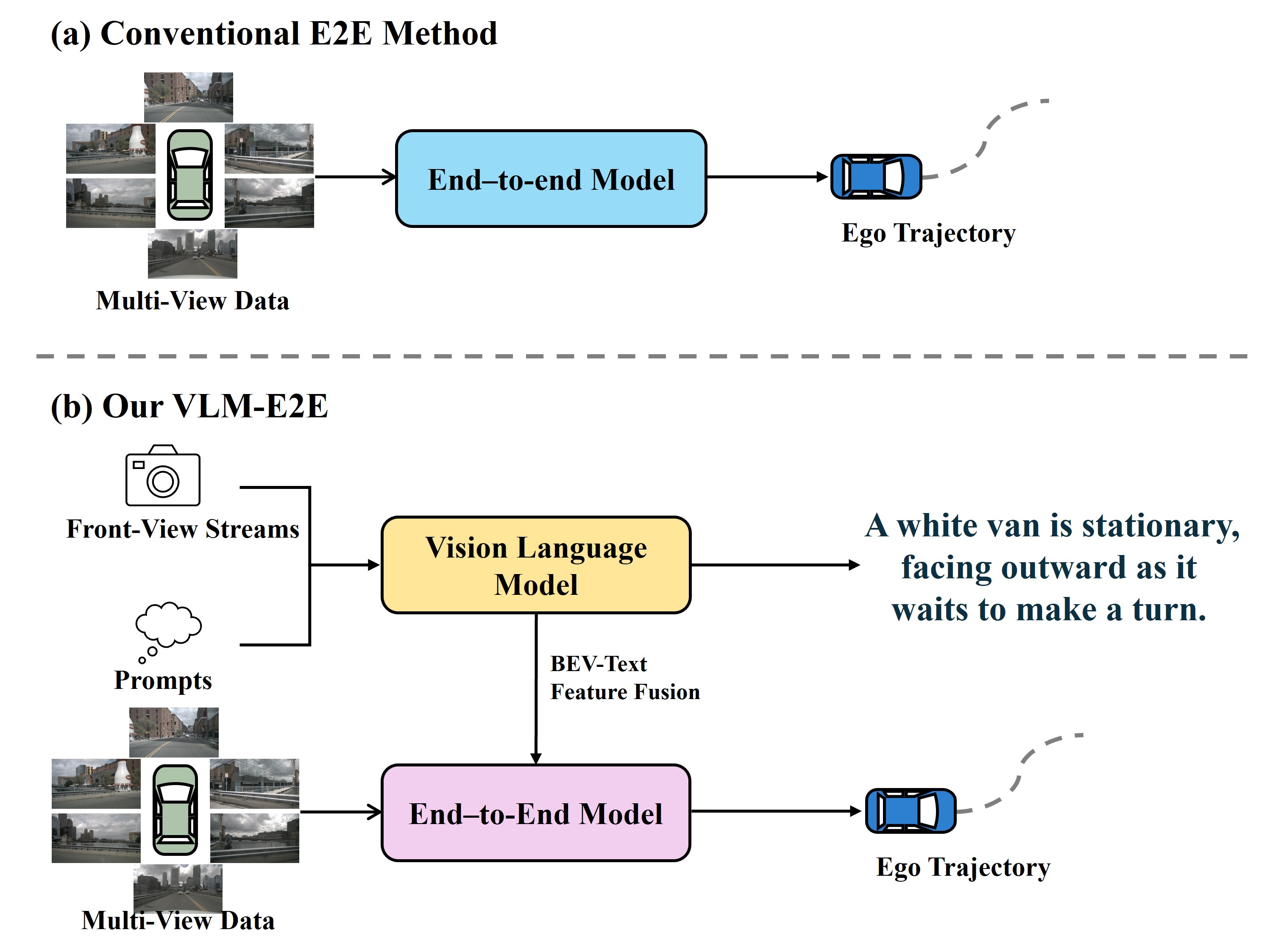}
    \caption{{\model} augments the end-to-end driving model with semantic textual descriptions during training. These descriptions extract driver attention from VLMs to encourage the model to learn richer attentional semantics.}
    \label{fig:fig1}
\end{figure}

This limitation has spurred the integration of Vision-Language Models (VLMs) \cite{wang2025cogvlm, chen2024internvl} into autonomous driving frameworks. Trained on vast multimodal datasets, VLMs excel at tasks requiring high-level semantic reasoning, such as interpreting complex scenes, predicting dynamic interactions, and generating contextual descriptions. Their ability to leverage commonsense knowledge makes them particularly well-suited for addressing challenges in autonomous driving, such as understanding traffic rules, identifying vulnerable road users, and making safe decisions in ambiguous scenarios. By generating text-based descriptions of critical driving cues, VLMs can explicitly capture and prioritize regions of interest that align with human driver attention. This capability enables more human-like decision-making, particularly in safety-critical scenarios where attentional focus is paramount.

Motivated by these challenges, we propose {\model} (illustrated in Fig. \ref{fig:fig1}), a novel framework designed to enhance autonomous driving systems by incorporating a deeper understanding of driver attentional semantics. Our approach addresses three key questions:

\textbf{How to integrate VLMs with E2E models?}
While most existing methods integrate VLMs with decision-making modules \cite{xu2024vlm} or other high-level components \cite{ma2025leapvad}, leveraging their semantic understanding capabilities to enhance decision processes, our approach introduces a novel integration strategy. Instead of limiting VLMs to decision modules, we combine them directly with the BEV module, which is widely used to represent and process spatial information from multiple perspectives in autonomous driving. By integrating VLMs into the BEV module, we enable BEV representations to incorporate both visual and textual features, resulting in richer and more semantic-aware spatial understanding. This integration allows the model to not only perceive geometric structures but also reason about high-level driver attentional semantics.

\textbf{How to fuse vision and text representations?}
Existing methods for fusing vision and text representations predominantly rely on attention-based mechanisms \cite{vaswani2017attention, han2022survey}, such as cross-attention or co-attention modules, to align and enhance interactions between modalities. While effective, these approaches often rely on predefined attention mechanisms that lack flexibility in adapting to varying task requirements and are time and memory-consuming. To address this issue, we propose a BEV-Text learnable weighted fusion strategy, where the importance of each modality is dynamically determined through a learnable weight mechanism. This approach allows the model to adaptively emphasize visual or textual features based on their relevance to the task, leading to a more robust and context-aware multimodal representation. For instance, in scenarios requiring precise localization such as lane keeping, the model can prioritize BEV features, while in scenarios requiring high-level reasoning, such as red lights, it can emphasize text features.

\textbf{How to represent driver attentional environment?}
To effectively model a driver-attentional environment, we propose a multimodal framework that leverages vision-language representations. First, we utilize front-view images captured from the driving scene as input to BLIP-2 \cite{li2023blip} to generate initial textual descriptions of the environment. These descriptions provide a semantic understanding of key objects and events within the driver's vision scope. To address the hallucination problem of VLMs, we further refine these textual representations using ground truth annotations and high-level maneuvering intentions. This refinement ensures that the generated text is not only accurate but also contextually aligned with the driving task. Finally, the refined text is encoded into a dense representation using a pre-trained CLIP \cite{radford2021learning} model, which aligns the textual information with visual features in a shared embedding space. This textual representation enables the model to capture driver attentional cues, such as focusing on pedestrians near crosswalks or traffic signals at intersections, leading to more human-like decision-making and better safety performance.
The key contributions of this work can be summarized as follows:
\begin{itemize} 
\item We propose {\model}, a novel framework that leverages VLMs to enrich the training process with attentional understanding. By integrating semantic and contextual information, {\model} explicitly captures driver attentional semantics, which enables more human-like decision-making in complex driving scenarios.

\item We introduce a BEV-Text learnable weighted fusion strategy that dynamically balances the contributions of BEV and textual modalities. This adaptive fusion mechanism is computationally efficient, which requires minimal additional overhead while significantly enhancing the model’s adaptability and robustness. 

\item To address the hallucination problem of VLMs, we incorporate semantic refinement of textual annotations generated from front-view images. By leveraging ground truth (GT) labels and high-level maneuvering intentions, we ensure that the textual representations are both accurate and highly relevant to the driving task, enhancing the model’s ability to reason about critical driving cues.

\item Extensive experiments on the {\nuScenes} dataset demonstrate the superiority of {\model} over existing methods. Our framework achieves significant improvements in handling complex driving scenarios, showcasing its ability to integrate geometric precision with high-level semantic reasoning for safer and more reliable autonomous driving.
\end{itemize}
\section{Related Work}
\label{sec:formatting}

\begin{figure*}
    \centering
    \includegraphics[width=0.85\linewidth]{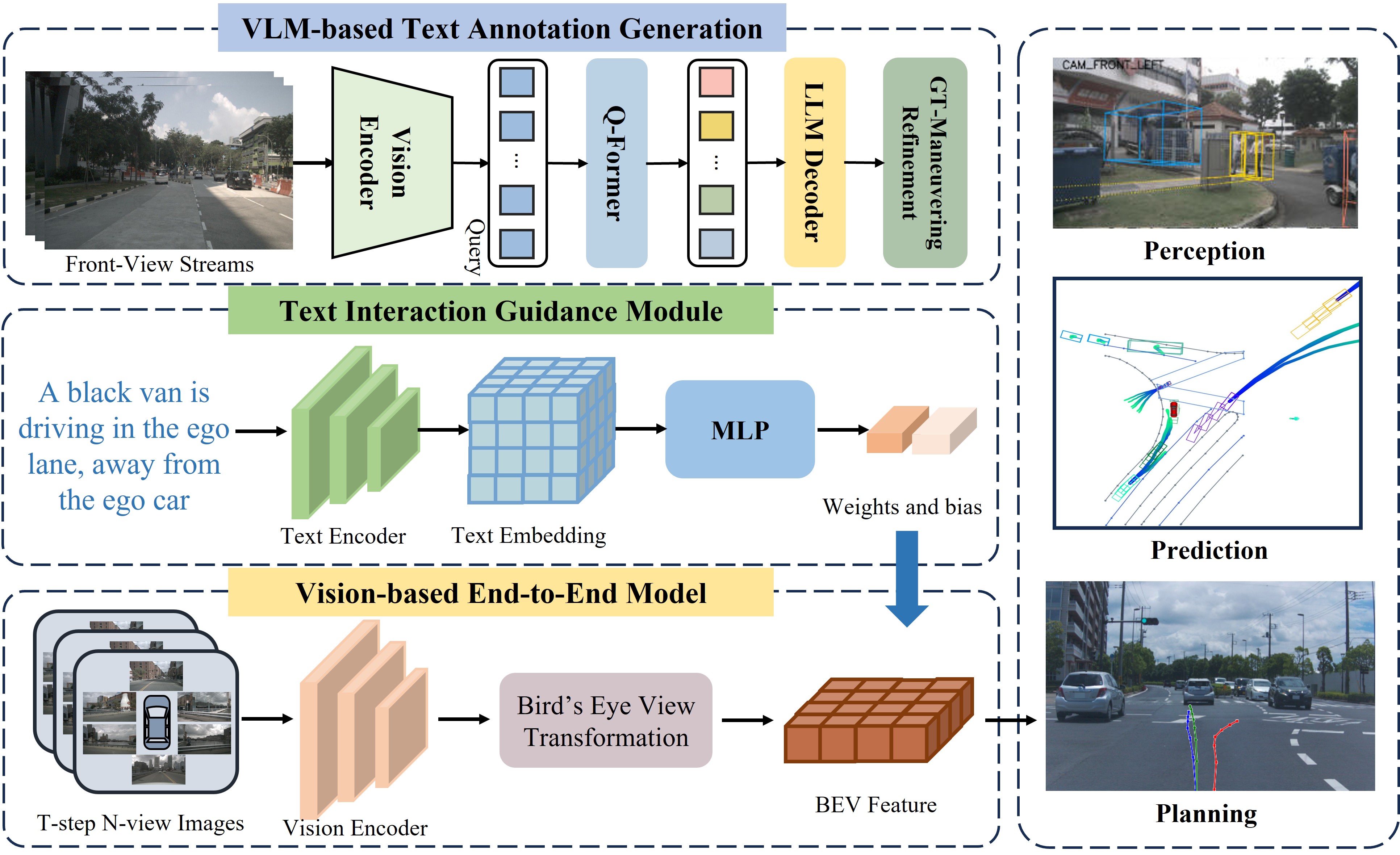}
    \caption{We present {\model}, a driver attention enhanced end-to-end vision-based framework. {\model} consists of three modules: VLM-based Text Annotation Generation, Text Interaction Guidance Module, and Vision-based End-to-end Model.}
    \label{fig:architect}
\end{figure*}

\subsection{End-to-End Autonomous Driving} 
Recent advances in end-to-end autonomous driving systems have established vision-based frameworks like ST-P3 \cite{hu2022st} and UniAD \cite{hu2023planning}, which unify perception, prediction, and planning for improved scene understanding. Follow-up works such as VAD \cite{jiang2023vad} and VADv2 \cite{chen2024vadv2} further enhanced dynamic environment handling through vectorized scene representations. Subsequent innovations like Ego-MLP \cite{zhai2023rethinking} and BEV-Planner \cite{li2024ego} introduce modular designs focusing on ego-state modeling and environment interactions.


\subsection{Vision Language Models in Autonomous Driving}
VLMs have gained traction in autonomous driving for their ability to enhance reasoning and interpretability in end-to-end systems. Recent studies demonstrate diverse integration approaches. For instance, Drive-with-LLMs \cite{chen2024driving} processes perception data via transformers for trajectory prediction, while DriveGPT4 \cite{xu2024drivegpt4} generates control signals with natural language explanations. Frameworks like DriveMLM \cite{wang2023drivemlm} and ELM \cite{zhou2024embodied} validate VLM-based planning through simulation and cross-domain pretraining.

Specialized datasets \cite{sima2024drivelm, qian2024nuscenes} and hybrid systems such as DriveVLM \cite{tian2024drivevlm}, which refines VLM outputs into precise trajectories, and VLM-AD \cite{xu2024vlm} and Senna \cite{jiang2024senna}, which simplify decision-making through text-based planning, highlight the field's advancements. However, existing methods rarely address the design of driving-specific textual semantics, particularly those capturing driver-attentional cues essential for human-like situational awareness. Our work bridges this gap by integrating attentional textual prompts to prioritize safety-critical scenarios.

\section{Methodology}
\label{sec:method}

In this section, we provide a detailed introduction to VLM-E2E, as illustrated in Fig. \ref{fig:architect}. The input scene information includes multi-view image sequences, GT, maneuvering, and user prompts. The front-view image, maneuvering, and user prompts are fed into the VLM-based Text Annotation Generation module to generate descriptive text annotations, while the multi-view images are processed by the visual encoding layer to produce BEV features. These text annotations are then passed to the Text Interaction Guidance Module, where they are encoded into text features using a pre-trained CLIP model. Subsequently, the BEV and text features are fused to support downstream tasks such as perception, prediction, and decision-making. In Section \ref{sec:vlm}, we introduce the design of VLM-based Text Annotation Generation in detail. Sections \ref{sec:text-e2e} and \ref{sec:e2e} focus on the design of the Text Interaction Guidance Module and Vision-based End-to-End Architecture, respectively.

\subsection{VLM-based Text Annotation Generation}
\label{sec:vlm}
\subsubsection{Text Annotation}
Fig. \ref{fig:architect} depicts the proposed pipeline for extracting driver attentional information from visual inputs, leveraging the reasoning capabilities of a pre-trained VLM. The semantic annotation extraction process can be formulated as follows:

\begin{equation}
    T = \mathcal{BLIP}_{2}(P, I_{front})
\end{equation}
where $\mathcal{BLIP}_{2}(\cdot)$ denotes the visual language model BLIP-2, $P$ represents the task-specific prompts, $I_{front}$ is the visual input from the ego vehicle's front camera, and $T$ is the generated textual description providing detailed environment-related information. The goal of this process is to utilize task-specific prompts alongside real-time visual inputs to extract actionable and attentional information from BLIP-2. This approach not only emphasizes critical elements such as pedestrians, traffic signals, and dynamic obstacles but also filters out irrelevant scene details, ensuring that the outputs directly support driving decisions.

In our work, we employ a state-of-the-art vision language model BLIP-2 \cite{li2023blip}, capable of performing complex reasoning over visual contexts, to generate precise and contextually relevant descriptions. The model interprets visual scenarios guided by prompts and outputs textual descriptions. This method enhances the dataset's richness by providing driver attentional annotations, thereby improving the understanding and decision-making capabilities of downstream driving models.

We encountered a challenge in determining the visual input. That is, selecting the right images from multiple cameras that can cover 360 degrees of the ego vehicle. Considering that we want to capture the driver’s attentional semantics when driving, the front view images usually contain the most relevant information required for most driving tasks. All-view images contain more distracting information that affects the system’s decision-making, so we choose to use only the front-view images to extract the attentional information. In addition, considering that the ego vehicle and its surroundings are in dynamic motion and the hallucination problem inherent in large models, we use the GT and maneuvering to
refine the annotations of dynamic objects, inspired by \cite{moon2024visiontrap}.

\subsection{Text Interaction Guidance Module}
\label{sec:text-e2e}
The driver's attentional text descriptions preserve rich visual semantic cues. It is complementary to the BEV features that mainly represent the 3D geometric information. Hence, BEV-Text fusion is proposed for comprehensive scene understanding from the BEV perspective. 

\subsubsection{Text Encoder}
Given a text input $T$ that provides semantic features to guide the BEV-Text fusion network toward achieving a specified fusion result, the text encoder and embedding within the text interaction guidance architecture are responsible for transforming this input into a text embedding. Among various VLMs, we adopt CLIP \cite{radford2021learning} due to its lightweight architecture and efficient text feature extraction capabilities. Compared to other VLMs, CLIP is computationally less demanding and produces text embeddings with relatively small text tokens of $77$, which significantly enhances the efficiency of subsequent BEV-Text feature fusion.
We freeze the text encoder from CLIP to preserve its consistency and leverage its pretrained knowledge. This process can be formally expressed as:
\begin{equation}
    f_{t} = \mathcal{CLIP}_{e}(T)
\end{equation}
where $\mathcal{CLIP} \in \mathbb{R}^{N \times L}$ denotes the CLIP model with wights frozen. $f_{t}$ is the text semantic representations.

In different but semantically similar texts, the extracted features should be close in the reduced Euclidean space. Furthermore, we use the MLP $F_{m}^i$ to mine this connection and further map the text semantic information and the semantic parameters. Therefore, it can be obtained:
\begin{equation}
    \gamma_m = F_m^1(f_{t}), \beta_m=F_m^2(f_{t})
\end{equation}
where $F_m^1$ and $F_m^2$ are the chunk operations to form the text semantic parameters.

As depicted in Fig. \ref{fig:architect}, in the text interaction guidance module, textual parameters interact through feature interaction and BEV features $s_t$, to obtain the effect of guidance. The feature modulation consists of scale scaling and bias control, which adjust the features from two perspectives, respectively. In particular, a residual connection is used to reduce the difficulty of network fitting, inspired by \cite{yi2024text}. For simplicity, it can be described as:
\begin{equation}
    x_t = (1 + \gamma_m) \odot s_t + \beta_m
\end{equation}
where $\odot$ denotes the Hadamard product. $x_t$ denotes the fused BEV feature, and $s_t$ denotes the BEV feature defined in Section \ref{subsec:bev}.

\subsection{Vision-based End-to-End Model}
\label{sec:e2e}
\subsubsection{Spatial Temporal BEV Perception}
\label{subsec:bev}
In our framework, the BEV representation is constructed from multi-camera images. The input multi-camera images $\{I_t^1, \cdots, I_t^n\}, n = 6$ at time $t$ are first passed through a shared backbone network, EfficientNet-b4 \cite{tan1905rethinking}, to extract high-dimensional feature maps. For each camera image $k$ at time $t$, we get its encoder features $e_t^k \in \mathbb{R}^{C \times H_e \times W_e}$ and depth estimation $d_t^k \in \mathbb{R}^{D \times H_e \times W_e}$ with $C$ denotes the number of feature channels, $D$ is the number of discrete depth values and $(H_e, W_e)$ depicts the spatial feature size. Implicit depth estimation is applied to infer the depth information for each pixel, enabling the construction of a 3D feature volume. Since the depth values are estimated, we take the outer product of the features with the depth estimation. 
\begin{equation}
    \hat{e}_t^k = e_t^k \otimes d_t^k
\end{equation}
where $\hat{e}_t^k \in \mathbb{R}^{C \times D \times H_e \times W_e}$. Then, to transform the 2D perspective features into a 3D space, we employ a feature lifting module. This module uses camera intrinsic and extrinsic parameters to project the 2D features into a 3D voxel space. The 3D feature volume is then collapsed into a 2D BEV representation by aggregating features along the vertical axis to form the BEV view features $b_t \in \mathbb{R}^{C \times H \times W}$, with $(H, W)$ denoting the spatial size of BEV feature. This is achieved through attention-based aggregation, which preserves the most salient features while maintaining spatial consistency. The resulting BEV map provides a top-down view of the scene, encapsulating both geometric and semantic information.

In addition to the BEV construction pipeline described above, we further incorporate temporal modeling to enhance the dynamic understanding of the scene. Specifically, given the current timestamp $t$ and its $h$ historical BEV features $\{b_{t-h}, \cdots, b_{t-1}, b_t\}$, we first align the historical features to the current frame's coordinate system using a temporal alignment module. This process leverages the relative transformation and rotation matrix $M_{t-i \to t} \in \mathbb{R}^{4 \times 4}$ between adjacent frames. The past BEV feature $b_{t-i}$ is then spatially transformed as: 
\begin{equation}
\hat{b}_{t-i} = \mathcal{W}(b_{t-i}, M_{t-i \to t}), \quad \ i=1,2
\end{equation}
where $\mathcal{W}(\cdot)$ denotes the pose-based BEV feature warping operation, and $\hat{b}_{t-i}$ represents the aligned historical features. Subsequently, the aligned BEV features from the $h$ frames are concatenated to form the spatiotemporal input $\hat{b}=[\hat{b}_{t-h}, \cdots, \hat{b}_{t-1}, \hat{b}_t] \in \mathbb{R}^{h \times C \times H \times W}$. To capture long-term dependencies in dynamic scenes, we use a spatiotemporal transformer module $F_s$. 
\begin{equation}
    s_t = F_s(\hat{b}_{t-h}, \cdots, \hat{b}_{t-1}, \hat{b}_t)
\end{equation}
where $s_t \in \mathbb{R}^{h \times C \times H \times W}$ is the spatiotemporally fused BEV feature. $F_s$ is a spatiotemporal convolutional unit with cross-frame self-attention. Our spatial-temporal BEV representation explicitly models the static and dynamic evolution of the scene, enabling the BEV representation to encode geometric structures and temporal continuity simultaneously.

\subsection{Attention Guided Future Planning}
The primary objective of the proposed motion planning is to generate trajectories that ensure safety, comfort, and efficient progress toward the goal. To achieve this, we employ a motion planner. The planner generates a set of kinematically feasible trajectories, each of which is evaluated using a learned scoring function, inspired by \cite{casas2021mp3, sadat2020perceive, zeng2019end, hu2022st}.

To ensure real-time performance, the set of sampled trajectories must remain sufficiently small. However, this set must also represent various possible maneuvers and actions to avoid encroaching obstacles. To strike this balance, we employ a sampling strategy that is aware of the lane structure, ensuring that the sampled trajectories effectively capture a diverse range of driving behaviors while remaining computationally feasible.

In particular, we follow the trajectory sampling method proposed in \cite{sadat2019jointly, werling2010optimal}, where trajectories are generated by combining longitudinal motion with lateral deviations relative to specific lanes, such as the current SDV lane or adjacent lanes. This approach allows the planner to sample trajectories that adhere to lane-based driving principles while incorporating variations in lateral motion. These variations enable the motion planner to handle a wide array of traffic scenarios.

To ensure the planned trajectory adheres to driver attention on traffic regulations and routes, we utilize a temporal refinement module that dynamically integrates traffic regulations. Leveraging front-view camera features $e_{front}$ from the encoder, we initialize a GRU-based refinement network to iteratively adjust the initially selected trajectory. The front-view features explicitly encode traffic regulations semantics, enabling the model to halt at red lights or proceed through green signals. 
The recurrent architecture ensures smooth transitions between trajectory points, mitigating abrupt steering or acceleration changes.
\section{Experiments}
\subsection{Experimental Settings}
\textbf{Dataset.} We evaluate our method on the nuScenes dataset \cite{caesar2020nuscenes}, a large-scale autonomous driving benchmark comprising 1,000 diverse driving scenes, each spanning 20 seconds with annotations provided at 2\,Hz. The dataset features a 360° multi-camera rig comprising six synchronized cameras with minimal field-of-view overlap. Precise camera intrinsic and extrinsic are provided for each frame to ensure accurate spatial alignment.

All labels are transformed into the ego vehicle’s reference frame using GT future ego-motion, ensuring temporal consistency across frames. Besides, maneuvering is used to correct annotations from VLMs. 

\noindent\textbf{Implementation Details.} Our framework processes 1 second of historical sensor data, equivalent to 3 frames at 2Hz, to predict trajectories over a 3\,s horizon spanning 6 future frames. Inputs include synchronized multi-camera RGB streams with $224\times480$ pixel resolution and BEV grid maps covering a 100m×100m area at 0.5m per pixel resolution. Training employs the Adam optimizer with a learning rate of $2.0 \times 10^{-3}$ for 20 epochs, using mixed precision on 4 NVIDIA A6000 GPUs and a batch size of 6.




\begin{table}
    \centering
    \caption{Perception results. We report the semantic segmentation IoU (\%) in BEV.}
    \scalebox{1.0}{
    \begin{tabular}{c|cccc}
    \toprule
    Method &  Drivable Area & Lane & Vehicle & Pedestrian\\
    \midrule
    VED \cite{lu2019monocular} &60.82 & 16.74 & 23.28 & 11.93 \\
    VPN \cite{pan2020cross} & 65.97 &17.05 &28.17 &10.26 \\
    PON \cite{roddick2020predicting} & 63.05 &17.19 &27.91 &13.93 \\
    Lift-Splat \cite{philion2020lift} & 72.23 &19.98 &31.22 &15.02 \\
    IVMP \cite{wang2021learning} & \textbf{74.70} &20.94 &34.03 &17.38 \\
    FIERY \cite{hu2021fiery}& 71.97 &33.58 &38.00 &17.15 \\
    ST-P3 \cite{hu2022st} & 74.38 &   38.47&  38.79& 14.06 \\
    \midrule
    \model & 74.69 & \textbf{39.33} & \textbf{39.08} & \textbf{17.49}\\
    \bottomrule
    \end{tabular}}
    
    \label{tab:Perception}
\end{table}

\begin{table}
    \centering
    \caption{Prediction results. We report the semantic and instance segmentations in BEV for 2s in the future.}
    \scalebox{1.0}{
    \begin{tabular}{c|cccc}
    \hline
    Method & IoU $\uparrow$ & PQ $\uparrow$ & SQ $\uparrow$ & RQ $\uparrow$ \\
    \hline
    FIERY \cite{hu2021fiery} &36.20 & 27.80 & - & -\\
    ST-P3 \cite{hu2022st} &36.89 & 29.10 & \textbf{69.77} & 41.71\\
    \model & \textbf{38.54} & \textbf{29.83} & 69.56  & \textbf{42.88}\\ 
    \hline
    \end{tabular}}
    \label{tab:Prediction}
\end{table}

\begin{table}[h]
    \centering
    \caption{Planning results. We report the L2 (m) and Collision Rate CR (\%) across 1s, 2s, 3s.}
    \scalebox{0.9}{
    \begin{tabular}{c|cccc|cccc}
    \hline
    \multirow{2}{*}{Method}  & \multicolumn{4}{c|}{L2 (m) $\downarrow$} & \multicolumn{4}{c}{CR (\%) $\downarrow$}\\
    \cline{2-9}
    ~  & 1s & 2s & 3s & Avg. & 1s & 2s & 3s & Avg.\\
    \hline
    Vanilla \cite{codevilla2019exploring} & 0.50 & 1.25 & 2.80&1.52 & 0.68 &0.98 &2.76 & 1.47\\
    NMP \cite{chen2020learning} & 0.61 & 1.44 & 3.18& 1.74& 0.66 & 0.90 & 2.34& 1.30\\
    Freespace \cite{prakash2021multi}& 0.56 & 1.27 & 3.08 & 1.64& 0.65 & 0.86 & 1.64& 1.05\\
    ST-P3 \cite{hu2022st} & 1.33 & 2.11 & 2.90 & 2.11 &0.23 & 0.62 & 1.27& 0.71\\

    UniAD \cite{hu2023planning} & 0.48 &0.96& 1.65 &1.03 &0.05 &0.17 &0.71 &0.31 \\
    VAD \cite{jiang2023vad} & 0.54& 1.15& 1.98& 1.22& 0.04 &0.39& 1.17 &0.53 \\
    GenAD \cite{zheng2024genad} & 0.28& 0.49& 0.78 & 0.52& 0.08& 0.14& 0.34 & 0.19\\
    Senna \cite{jiang2024senna}& 0.37 &0.54 &0.86& 0.59 &0.09 &0.12 &0.33 &0.18 \\
    VLM-E2E & \textbf{0.28} & \textbf{0.50} & \textbf{0.80} & \textbf{0.53}& \textbf{0.01} & \textbf{0.06} & \textbf{0.20} & \textbf{0.09}\\
    \hline
    \end{tabular}}
    \label{tab:Planning}
\end{table}

\subsection{Quantitative Results}
\noindent \textbf{Perception.} Table \ref{tab:Perception} presents the perception performance of various methods across four key categories: Drivable Area, Lane, Vehicle, and Pedestrian. Our proposed {\model} model demonstrates significant improvements over existing approaches, achieving the best results in three out of four categories. Specifically, {\model} outperforms ST-P3 in lane detection with a $2.24\%$ relative improvement, vehicle detection with an $0.75\%$ increase, and Pedestrian detection with a $24.40\%$ boost on the nuScenes validation set. While IVMP achieves the highest score in drivable area detection, {\model} closely follows with a score of 74.69, demonstrating preserved geometric reasoning in BEV space. These results demonstrate that integrating driver-attentional features enhances critical perception tasks without compromising foundational scene understanding.

\begin{figure*}
    \centering
    \includegraphics[width=0.9\linewidth]{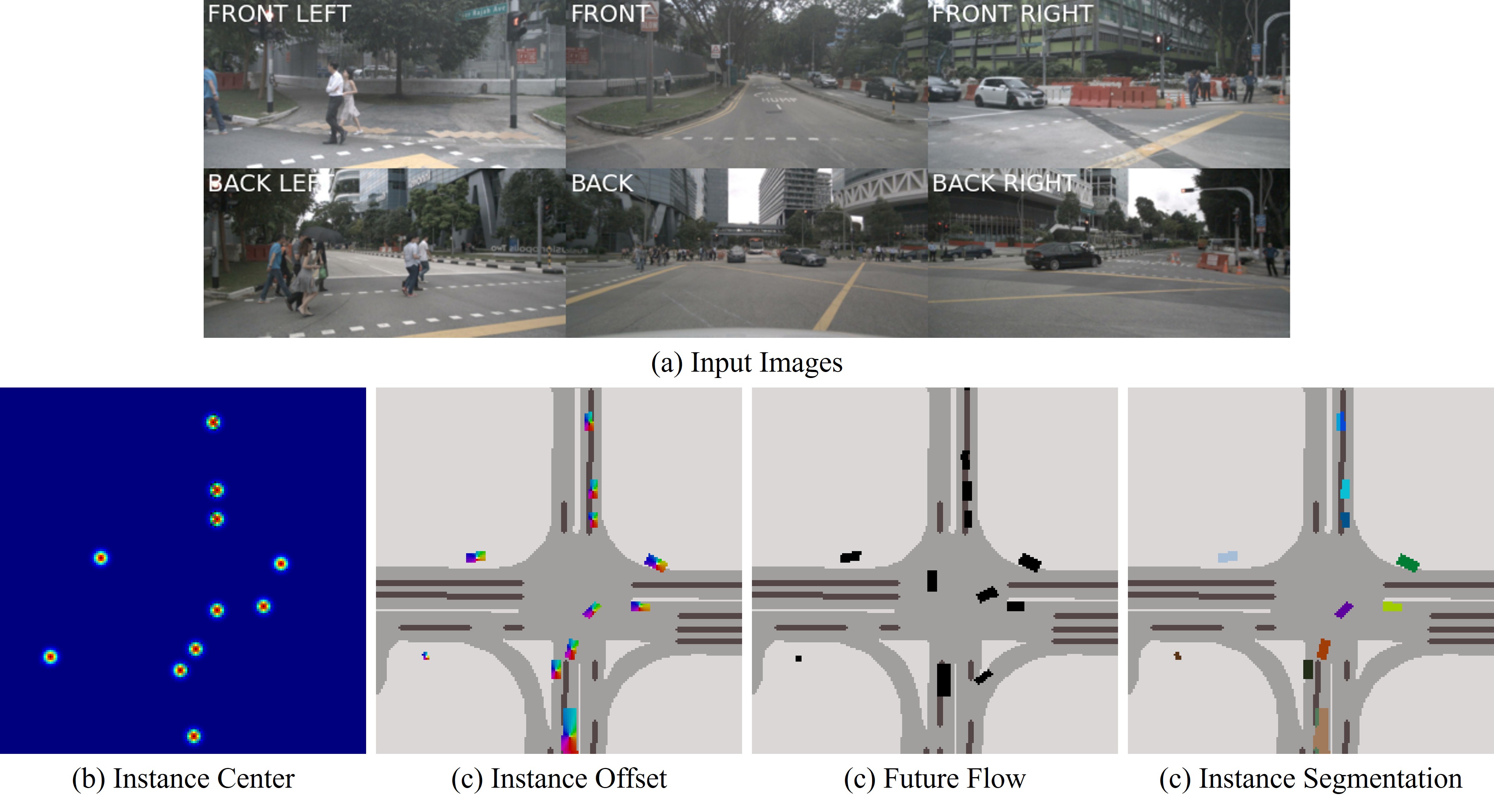}
    \caption{Qualitative analysis on prediction. (a) shows the multi-view input images. (b) shows the heatmap (blue to red) which illustrates the probability
distribution of instance centers within the scene, with warmer colors indicating higher confidence regions. (c) represents the vehicles segmentation, which
effectively distinguishes individual instances in the complex traffic scenario. (d) reveals the directional vectors pointing towards the corresponding instance
centers for each pixel, demonstrating the model’s understanding of spatial relationships. (e) exhibits consistency within each instance, reflecting the characteristic
rigid-body motion of vehicles.}
    \label{fig:fig8}
\end{figure*}


\noindent\textbf{Prediction.} Table \ref{tab:Prediction} presents the prediction performance of various methods on the task of semantic and instance segmentations in BEV for a future horizon of 2.0 seconds. We evaluate the methods using IoU, PQ, SQ, and RQ. {\model} achieves the best performance across IoU, Panoptic Quality (PQ), Segmentation Quality (SQ), and  Recognition Quality (RQ). Specifically, {\model} attains an IoU of 38.54, representing a $4.47\%$ improvement over ST-P3. In terms of PQ, {\model} achieves a PQ of 29.83, demonstrating superior instance segmentation capabilities compared to ST-P3 and FIERY. 
These results highlight the effectiveness of {\model} in capturing both semantic and instance-level information in dynamic driving scenarios.

\noindent\textbf{Planning.} As shown in Table \ref{tab:Planning}, our VLM-E2E achieves the best overall performance among all methods. It achieves the lowest average L2 error, 0.53\,m, and substantially reduces the collision rate to 0.09\%. Compared to previous state-of-the-art approaches, such as GenAD, which achieves 0.52 meters and 0.19 percent, and Senna, which reaches 0.59\,m and 0.18\%, VLM-E2E delivers notable improvements in both trajectory accuracy and safety. Moreover, the gains over traditional baselines like Vanilla and NMP are even more significant. These results clearly demonstrate the effectiveness of incorporating VLM-guided attention into end-to-end autonomous driving frameworks.

\begin{table}[]
    \centering
    \caption{Ablations of text encoders. We report the semantic segmentation
IoU (\%) in BEV.}
    \scalebox{1.0}{
    \begin{tabular}{c|cccc}
    \hline
     Method &  Drivable Area & Lane & Vehicle & Pedestrian\\
     \hline
     Bert &  73.44 & 38.19& 38.53 & 15.90\\
     CLIP &  \textbf{74.69} &\textbf{39.33} &\textbf{39.08} &\textbf{17.49}\\
    \hline
    \end{tabular}}
    \label{tab:ablation1}
\end{table}

\begin{table}[]
    \centering
    \caption{Ablations of input text views. We report the semantic segmentation IoU (\%) in BEV.}
    \scalebox{1.0}{
    \begin{tabular}{c|cccc}
    \hline
         Text View &  Drivable Area & Lane & Vehicle & Pedestrian\\
         \hline
         All View &73.50 &37.98&38.79&\textbf{17.51}\\
         Front View &  \textbf{74.69} &\textbf{39.33} &\textbf{39.08} &17.49\\
    \hline
    \end{tabular}}
    \label{tab:ablation2}
\end{table}

\subsection{Qualitative Analysis}

Fig. \ref{fig:fig8} demonstrates the generated outputs, including instance segmentation, instance center, instance offset, and future flow. Fig. 3(b) features a heatmap highlighting detected
objects, while Fig. 3(c) displays the instance segmentation results, where each segment is color-coded to represent different
objects. The offset is a vector pointing to the center of the
instance in Fig. 3(d). The future flow Fig. 3(e) is a displacement vector field of the dynamic agents. These visualizations
enhance the understanding of spatial relationships and the
distribution of elements within the environment, underscoring
the model’s capability to accurately perceive and segment
critical features essential for autonomous driving applications

\begin{figure*}
    \centering
    \includegraphics[width=0.9\linewidth]{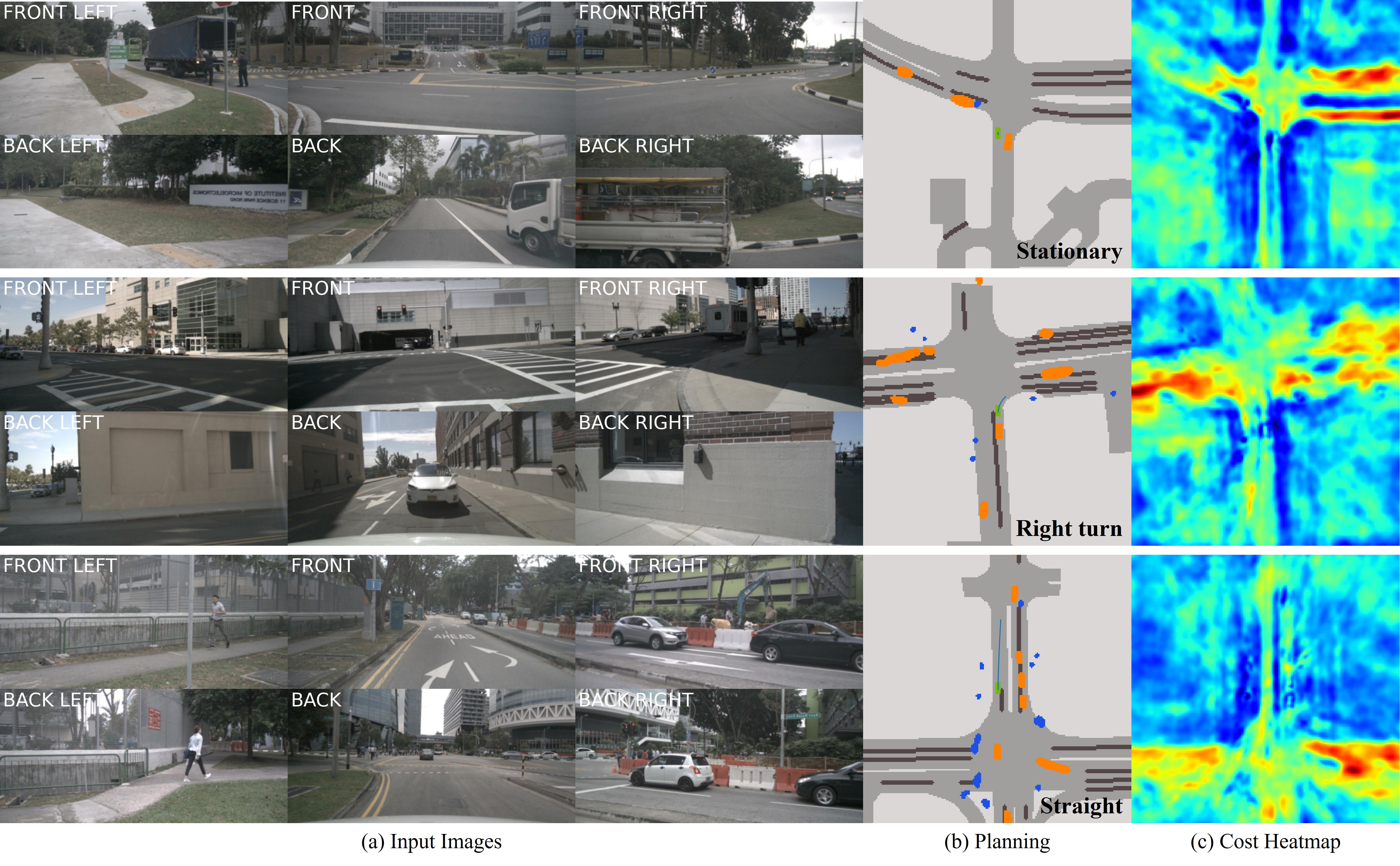}
    \caption{Qualitative analysis on planning. (a) shows the multi-view input images. (b) shows the planned trajectory (blue). (d) presents the learned costmap
with a warmer color indicates a lower cost.
}
    \label{fig:planning}
\end{figure*}

Fig. \ref{fig:planning} illustrates examples of planning scenarios. In the upper scene, the model accurately predicts the route when provided with turning instructions, effectively navigating through crowded environments in a manner similar to human demonstrations. The bottom scene demonstrates the model's predictions when instructed to proceed straight at an intersection, further highlighting its ability to handle diverse driving scenarios with precision. These examples emphasize the model's advanced planning capabilities in complex and dynamic environments.



\subsection{Ablation Study}
\textbf{Text Encoder.} Table \ref{tab:ablation1} compares the performance of text encoders, Bert \cite{devlin2019bert} and CLIP, across four detection categories. Bert shows a consistent performance decline, with decreases of 1.67\%, 2.90\%, 1.41\%, and 9.09\%, respectively, highlighting its limitations in perception tasks. In contrast, CLIP demonstrates superior capability in encoding textual annotations, achieving higher accuracy across all categories. The results validate that CLIP's joint embedding space effectively mitigates modality gaps between text annotations and BEV representations.

\noindent\textbf{Input Text View.} Table \ref{tab:ablation2} evaluates the impact of text view selection on BEV semantic segmentation performance. When transitioning from all view to front view text inputs, we observe performance improvements of 1.62\% for drivable area, 3.55\% for lane, and 0.75\% for vehicle, respectively. This improvement primarily stems from eliminating interference caused by redundant textual context in all views, where excessive multi-view descriptions introduce semantic noise that distracts the model from task-critical features. The results validate our design choice to adopt the front view as the default configuration, as it optimally balances semantic specificity and noise suppression for scene understanding tasks.

\section{Conclusion}
We propose {\model}, an E2E autonomous driving framework that leverages VLMs to enhance driver-attentional semantic understanding. We introduce a BEV-Text learnable weighted fusion strategy balancing geometric-semantic features to address modality imbalance and semantic underutilization in existing systems, along with a spatiotemporal coherence mechanism for temporal consistency, and an attention-guided trajectory refinement for probabilistic prediction. Future work will extend integration with advanced E2E architectures for improved generalization.

\bibliography{root}
\bibliographystyle{ieeetr}

\end{document}